\documentclass{article} %
\usepackage{iclr2020_conference,times}

\usepackage{amsmath,amsfonts,bm}

\def\eqref#1{equation~\ref{#1}}

\def\1{\bm{1}}

\DeclareMathAlphabet{\mathsfit}{\encodingdefault}{\sfdefault}{m}{sl}
\SetMathAlphabet{\mathsfit}{bold}{\encodingdefault}{\sfdefault}{bx}{n}

\def\sR{{\mathbb{R}}}
\def\sS{{\mathbb{S}}}

\usepackage{hyperref}
\usepackage{url}
\usepackage{color}
\usepackage{bbm}
\usepackage{graphicx}
\usepackage{tabularx}
\usepackage{xspace}
\usepackage{caption}
\usepackage{booktabs}
\usepackage{microtype}
\usepackage{amssymb}
\usepackage{wrapfig}

\newcommand{\nerf}{NeRF\xspace}
\newcommand{\nerfplusplus}{NeRF++\xspace}

\title{\nerfplusplus: Analyzing and Improving\\ Neural Radiance Fields}

\iclrfinalcopy

\author{
Kai Zhang\\
Cornell Tech
\And
Gernot Riegler\\
Intel Labs
\And
Noah Snavely\\
Cornell Tech
\And    
Vladlen Koltun\\
Intel Labs
}

\begin{document}

\maketitle

\begin{abstract}
Neural Radiance Fields (\nerf) achieve impressive view synthesis results for a variety of capture settings, including 360$^\circ$ capture of bounded scenes and forward-facing capture of bounded and unbounded scenes.  \nerf fits multi-layer perceptrons (MLPs) representing view-invariant opacity and view-dependent color volumes to a set of training images, and samples novel views based on volume rendering techniques.
In this technical report, we first remark on radiance fields and their potential ambiguities, namely the \emph{shape-radiance ambiguity}, and analyze \nerf's success in avoiding such ambiguities. Second, we address a parametrization issue involved in applying \nerf to 360$^\circ$ captures of objects within large-scale, unbounded 3D scenes. Our method improves view synthesis fidelity in this challenging scenario.
Code is available at \url{https://github.com/Kai-46/nerfplusplus}.

\end{abstract}
\section{Introduction}
\label{sec:intro}

Recall your last vacation where you captured a few photos of your favorite place.
Now at home you wish to walk around in this special place again, if only virtually. 
This requires you to render the same scene from different, freely placed viewpoints in a possibly unbounded scene.
This \emph{novel view synthesis} task is a long-standing problem in computer vision and graphics \citep{Chen1993View, Debevec1996Modeling, levoy1996light, gortler1996lumigraph, Shum2000review}.

Recently, learning-based methods have led to significant progress towards photo-realistic novel view synthesis.
The method of Neural Radiance Fields (\nerf), in particular, has attracted significant attention~\citep{mildenhall2020nerf}. \nerf is an implicit MLP-based model that maps 5D vectors---3D coordinates plus 2D viewing directions---to opacity and color values, computed by fitting the model to a set of training views. 
The resulting 5D function can then be used to generate novel views with conventional volume rendering techniques.

In this technical report, we first present an analysis of potential failure modes in \nerf, and an analysis of why \nerf avoids these failure modes in practice. Second, we present a novel spatial parameterization scheme that we call \emph{inverted sphere parameterization} that allows \nerf to work on a new class of captures of unbounded scenes.

\begin{figure}[t]
\centering
    \begin{tabular}{c@{\hspace{3em}} c@{\hspace{3em}}} \\
        \includegraphics[width=0.35\textwidth]{{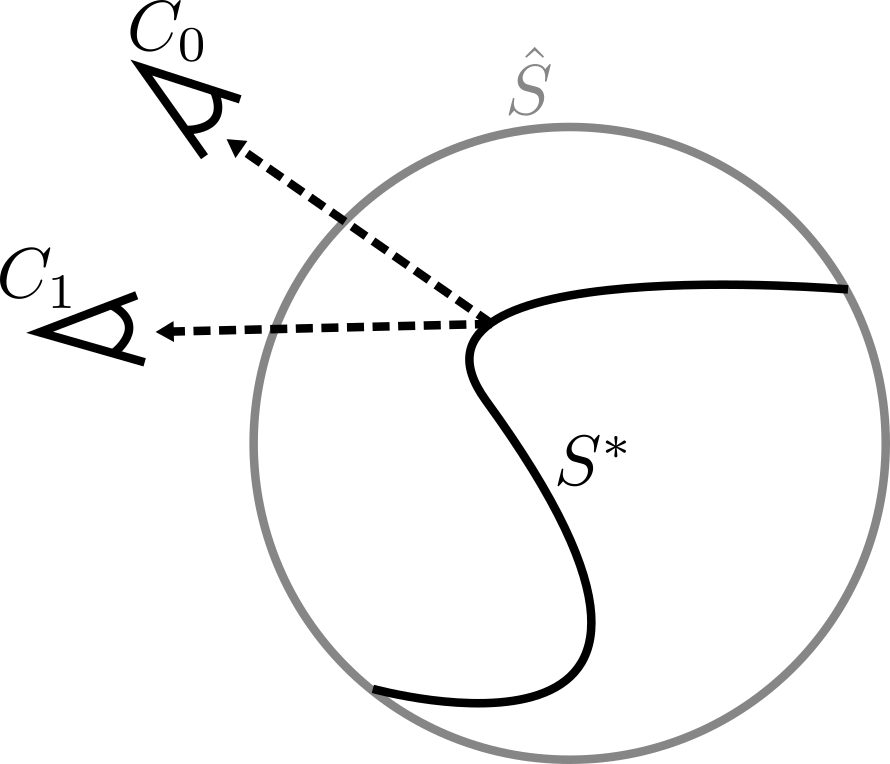}}  & 
        \includegraphics[width=0.35\textwidth]{{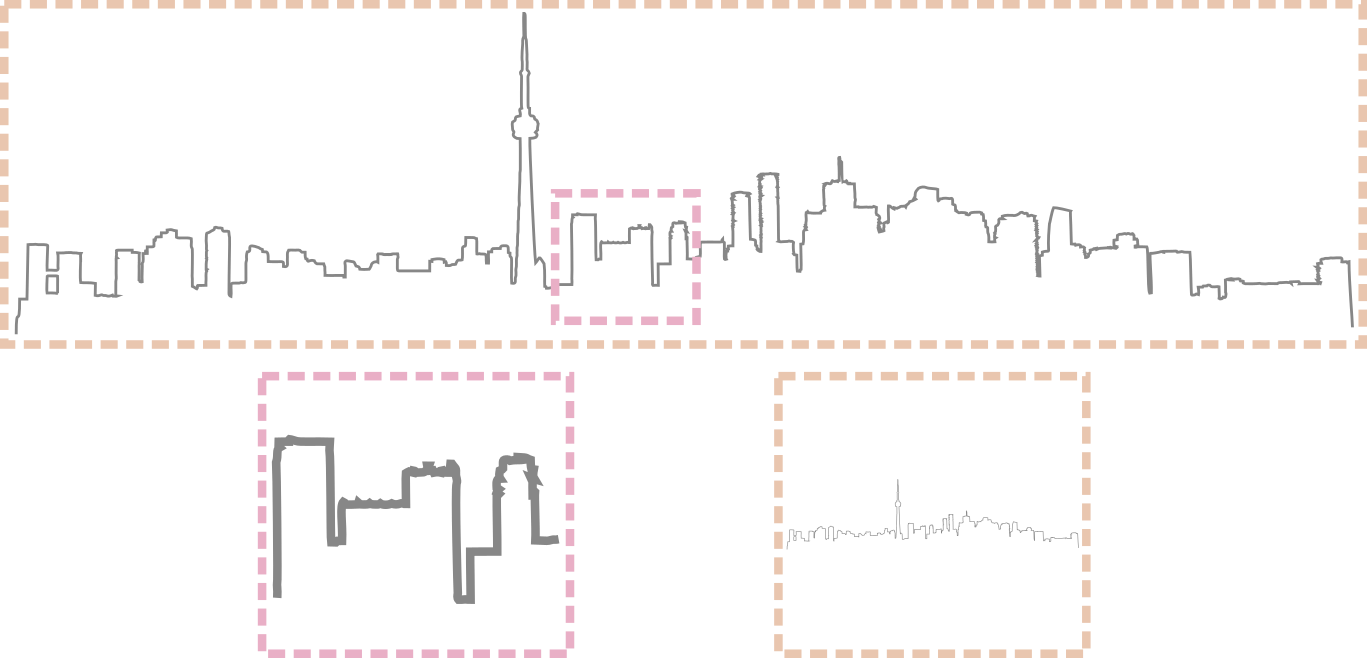}} \\
    {\small Shape-Radiance Ambiguity}  & {\small Parameterization of Unbounded Scenes}
     \end{tabular} 
  \caption{
  \emph{Shape-radiance ambiguity (left) and parameterization of unbounded scenes (right).} \textbf{Shape-radiance ambiguity:} our theoretical analysis shows that, in the absence of explicit or implicit regularization,
    a set of training images can be fit independently of the recovered geometry (e.g., for incorrect scene geometry $\hat{S}$ rather than correct geometry $S^*$) by exploiting view-dependent radiance to simulate the effect of the correct geometry. \textbf{Parameterization of unbounded scenes:} with standard parameterization schemes, either (1) only a portion of the scene is modeled (red outline), leading to significant artifacts in background elements, or (2) the full scene is modeled (orange outline), which leads to an overall loss of details due to finite sampling resolution.
  }
  \label{fig:illustration_of_all}
\end{figure}

In particular, we find that in theory, optimizing the 5D function from a set of training images can encounter critical degenerate solutions that fail to generalize to novel test views, in the absence of any regularization. Such phenomena are encapsulated in the \emph{shape-radiance ambiguity} (Figure~\ref{fig:illustration_of_all}, left), wherein one can fit a set of training images perfectly for an arbitrary incorrect geometry by a suitable choice of outgoing 2D radiance at each surface point. We empirically show that the specific MLP structure used in \nerf plays an important role in avoiding such ambiguities, yielding an impressive ability to synthesize novel views. Our analysis offers a new view into \nerf's impressive success. 

We also address a spatial parameterization issue that arises in challenging scenarios involving 360$^\circ$ captures around objects within unbounded environments (Figure~\ref{fig:illustration_of_all}, right). 
For 360$^\circ$ captures, \nerf assumes that the entire scene can be packed into a bounded volume, which 
is problematic for large-scale scenes: either we fit a small part of the scene into the volume and sample it in detail, but completely fail to capture background elements; or, we fit the full scene into the volume and lack detail everywhere due to limited sampling resolution. We propose a simple yet effective solution that separately models foreground and background, addressing the challenge of modeling unbounded 3D background content with an inverted sphere scene parameterization. We show improved quantitative and qualitative results on real-world captures from the Tanks and Temples dataset~\citep{Knapitsch2017} and from the light field dataset of \citet{yucer2016efficient}. 

In summary, we present an analysis on how \nerf manages to resolve the shape-radiance ambiguity, as well as a remedy for the parameterization of unbounded scenes in the case of 360$^\circ$ captures.

\section{Preliminaries}
Given posed multi-view images of a static scene, \nerf reconstructs an opacity field $\sigma$ representing a soft shape,
along with a radiance field $\mathbf{c}$ representing view-dependent surface texture. Both $\sigma$ and $\mathbf{c}$ are represented implicitly as multi-layer perceptrons (MLPs); the opacity field is computed as a function of 3D position, $\mathbf{x}\in\sR^3$, and the radiance field is parameterized by both 3D position and viewing direction, $\mathbf{d}\in \sS^2$ (i.e., the set of unit 3-vectors). Hence, we use $\sigma(\mathbf{x})$ to refer to opacity as a function of position, and $\mathbf{c}(\mathbf{x}, \mathbf{d})$ to refer to radiance as a function of position and viewing direction.

Ideally, $\sigma$ should peak at the ground-truth surface location for opaque materials, in which case $\mathbf{c}$ reduces to the surface light field~\citep{wood2000surface}. 
Given $n$ training images, \nerf uses stochastic gradient descent to optimize $\sigma$ and $\mathbf{c}$ by minimizing the discrepancy between the ground truth observed images $I_i$, and the predicted images $\hat{I}_i(\sigma, \mathbf{c})$ rendered from $\sigma$ and $\mathbf{c}$ at the same viewpoints:
\begin{align}
\label{eq:nerf_objective}
    \min_{\sigma, \mathbf{c}} \frac{1}{n} \sum_{i=1}^n \lVert I_i - \hat{I}_i(\sigma, \mathbf{c})\rVert ^2_2 \,.
\end{align}
The implicit volumes $\sigma$ and $\mathbf{c}$ are ray-traced to render each pixel of $\hat{I}(\sigma, \mathbf{c})$~\citep{kajiya1984ray}.
For a given ray $\mathbf{r} = \mathbf{o} + t\mathbf{d}$, $\mathbf{o} \in \sR^3, \mathbf{d}\in S^2, t\in\sR^+$, its color is determined by the integral
\begin{align}
    \mathbf{C}(\mathbf{r})=\int_{t=0}^{\infty} \sigma(\mathbf{o}+t\mathbf{d})\cdot \mathbf{c}(\mathbf{o}+t\mathbf{d}, \mathbf{d})\cdot e^{-\int_{s=0}^{t}\sigma(\mathbf{o}+s\mathbf{d})ds}dt \,.
\label{eq:integral}
\end{align}
To compensate for the network's spectral bias and to synthesize sharper images, \nerf uses a positional encoding $\gamma$ that maps $\mathbf{x}$ and $\mathbf{d}$ to their Fourier features~\citep{tancik2020fourier}:
\begin{align}
\label{eq:positional_encoding}
\gamma^{k}: \mathbf{p} \xrightarrow[]{}\big(\sin(2^0\mathbf{p}), \cos(2^0\mathbf{p}),\sin(2^1\mathbf{p}), \cos(2^1\mathbf{p}), \dots, \sin(2^{k}\mathbf{p}), \cos(2^{k}\mathbf{p})\big) \,,
\end{align}
where 
$k$ is a hyper-parameter specifying the dimensionality of the Fourier feature vector.

\vspace{-3em}
\section{Shape-radiance ambiguity}\label{sec:analyzing}

The capacity of \nerf to model view-dependent appearance leads to an inherent ambiguity between 3D shape and radiance that can admit degenerate solutions, in the absence of regularization.
For an arbitrary, incorrect shape, one can show that there exists a family of radiance fields that perfectly explains the training images, but that generalizes poorly to novel test views.

\begin{figure}[t]
\centering
    \begin{tabular}{c@{\hspace{0.02em}}c@{\hspace{0.02em}}c@{\hspace{0.02em}}c@{\hspace{0.02em}}c@{\hspace{0.02em}}}
        \includegraphics[trim=0 60 0 40,clip,width=0.23\columnwidth]{{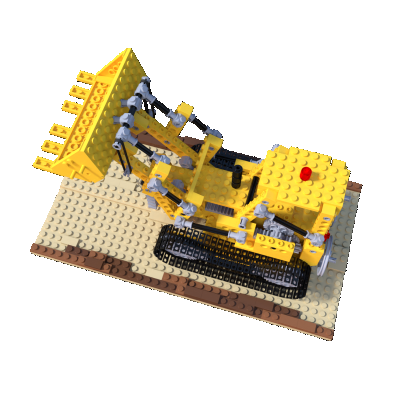}}  & 
        \includegraphics[trim=0 60 0 40,clip,width=0.23\columnwidth]{{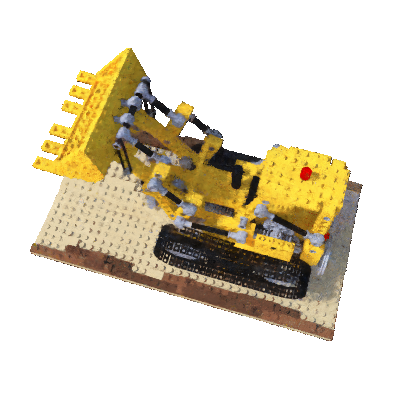}} & 
        \includegraphics[trim=0 60 0 40,clip,width=0.23\columnwidth]{{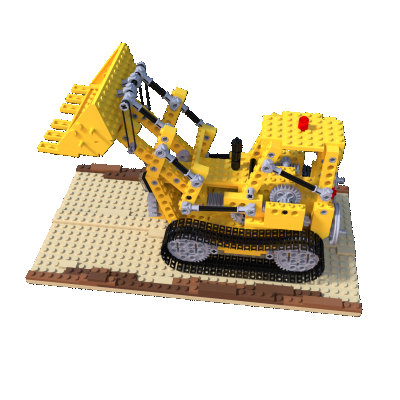}} &
        \includegraphics[trim=0 60 0 40,clip,width=0.23\columnwidth]{{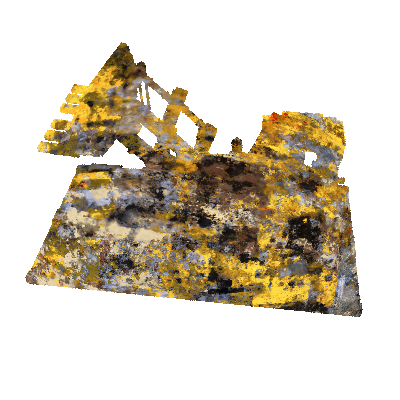}} \\
 {\small GT training view} & {\small Prediction (PSNR: 25.4)} & {\small GT test view} & {\small Prediction (PSNR: 10.7)}
     \end{tabular} 
  \caption{To demonstrate the shape-radiance ambiguity, we pretrain \nerf on a synthetic dataset where the opacity field $\sigma$ is optimized to model an incorrect 3D shape (a unit sphere, instead of a bulldozer shape), while the radiance field $\mathbf{c}$ is optimized to map the training rays' intersection with the sphere and view directions to their pixel color. In this example, we use 3 MLP layers 
  to model the effects of view-dependence (see the MLP structure in Figure~\ref{fig:mlp_structure}), and fit to 50 synthetic training images with viewpoints randomly distributed on a hemisphere. 
  The resulting incorrect solution explains the training images very well (left two images), but fails to generalize to novel test views (right two images).
  }\label{fig:existence_shape_radiance}
\end{figure}

To illustrate this ambiguity, imagine that for a given scene we represent the geometry as a unit sphere. 
In other words, let us fix \nerf's opacity field to be 1 at the surface of the unit sphere, and 0 elsewhere.
Then, for each pixel in each training image, we intersect a ray through that pixel with the sphere, and define the radiance value at the intersection point (and along the ray direction) to be the color of that pixel. This artificially constructed solution is a valid \nerf reconstruction that perfectly 
fits the input images. 
However, this solution's ability to synthesize novel views is very limited: accurately generating such a view would involve reconstructing an arbitrarily complex view-dependent function at each surface point. The model is very unlikely to accurately interpolate such a complex function, unless the training views are extremely dense, as in conventional light field rendering works~\citep{buehler2001unstructured,levoy1996light,gortler1996lumigraph}. 
This shape-radiance ambiguity is illustrated in Figure~\ref{fig:existence_shape_radiance}.

Why does \nerf avoid such degenerate solutions? 
We hypothesize that two related factors come to \nerf's rescue: 
1) incorrect geometry forces the radiance field to have higher intrinsic complexity (i.e., much higher frequencies) while in contrast 
2) \nerf's specific MLP structure implicitly encodes a smooth BRDF prior on surface reflectance. 

\textbf{Factor 1}: As $\sigma$ deviates from the correct shape, $\mathbf{c}$ must in general become a high-frequency function with respect to $\mathbf{d}$ to reconstruct the input images. 
For the correct shape, the surface light field will generally be much smoother (in fact, constant for Lambertian materials).
The higher complexity required for incorrect shapes is more difficult to represent with a limited capacity MLP. 

\begin{wrapfigure}{r}{0.45\textwidth}
\vspace{-1em}
  \begin{center}
    \includegraphics[trim=0 0 0 0,clip,width=0.42\textwidth]{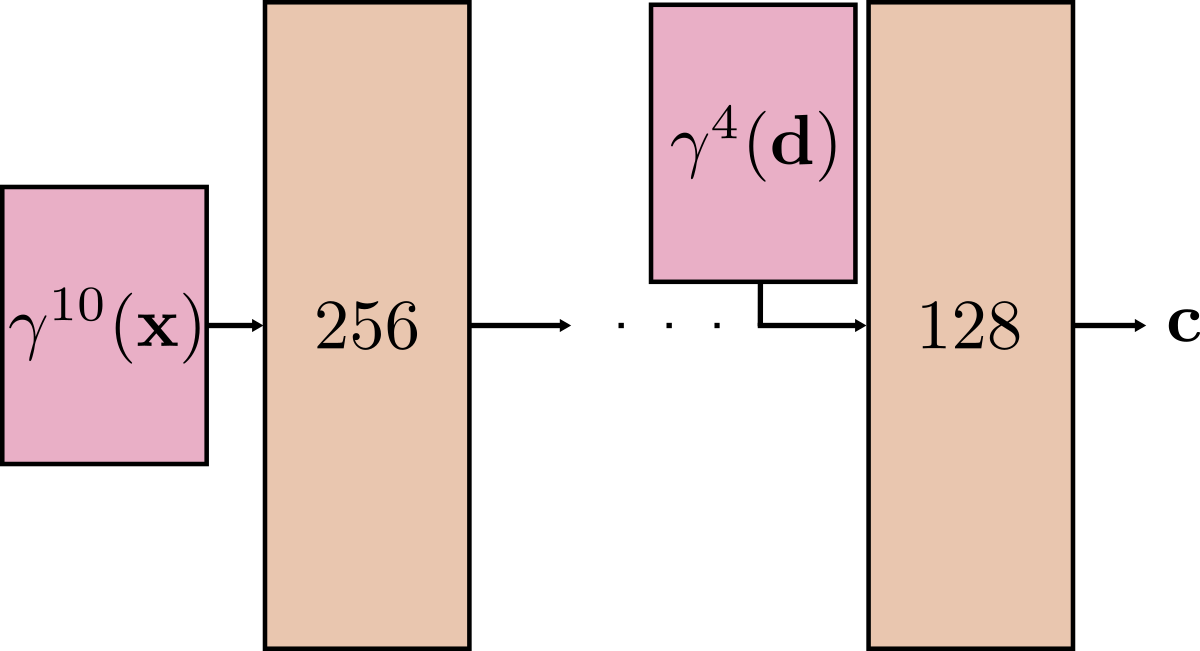}
  \end{center}
\vspace{-10pt}
  \caption{{\small Structure of \nerf's MLP for modeling radiance $\mathbf{c}$.}}
\vspace{-1em}
  \label{fig:mlp_structure}
\end{wrapfigure}

\textbf{Factor 2:} In particular, \nerf's specific MLP structure encodes an implicit prior favoring smooth surface reflectance functions where $\mathbf{c}$ is smooth with respect to $\mathbf{d}$ at any given surface point $\mathbf{x}$.
This MLP structure, shown in Figure~\ref{fig:mlp_structure}, treats the scene position $\mathbf{x}$ and the viewing direction $\mathbf{d}$ asymmetrically: 
$\mathbf{d}$ is injected into the network close to the end of the MLP, meaning that there are fewer MLP parameters, as well as fewer non-linear activations, involved in the creation of view-dependent effects. 
In addition, the Fourier features used to encode the viewing direction consist only of low-frequency components, i.e., $\gamma^4(\cdot)$ vs.\ $\gamma^{10}(\cdot)$ for encoding $\mathbf{d}$ vs.\ $\mathbf{x}$ (see Eq.~\ref{eq:positional_encoding}).
In other words, for a fixed $\mathbf{x}$, the radiance $\mathbf{c}(\mathbf{x},\mathbf{d})$ has limited expressivity with respect to $\mathbf{d}$.

To validate this hypothesis, we perform an experiment where we instead represent $\mathbf{c}$ with a vanilla MLP that treats $\mathbf{x}$ and $\mathbf{d}$ symmetrically---i.e., accepting both as inputs to the first layer and encoding both with $\gamma^{10}(\cdot)$---to eliminate any implicit priors involving viewing direction that arise from the network structure. 
If we train \nerf from scratch with this alternate model for $\mathbf{c}$, we observe reduced test image quality compared with \nerf's special MLP, as shown in  Figure~\ref{fig:implicit_regularize_nerf} and Table~\ref{tab:auxiliary_loss}. This result is consistent with our hypothesis that implicit regularization of reflectance in \nerf's MLP model of radiance $\mathbf{c}$ helps recover correct solutions.

\begin{minipage}{\textwidth}
\centering
    \begin{tabular}{c@{\hspace{0.02em}}c@{\hspace{0.02em}}c@{\hspace{0.02em}}}
        \includegraphics[trim=100 0 0 0,clip,width=0.33\columnwidth]{{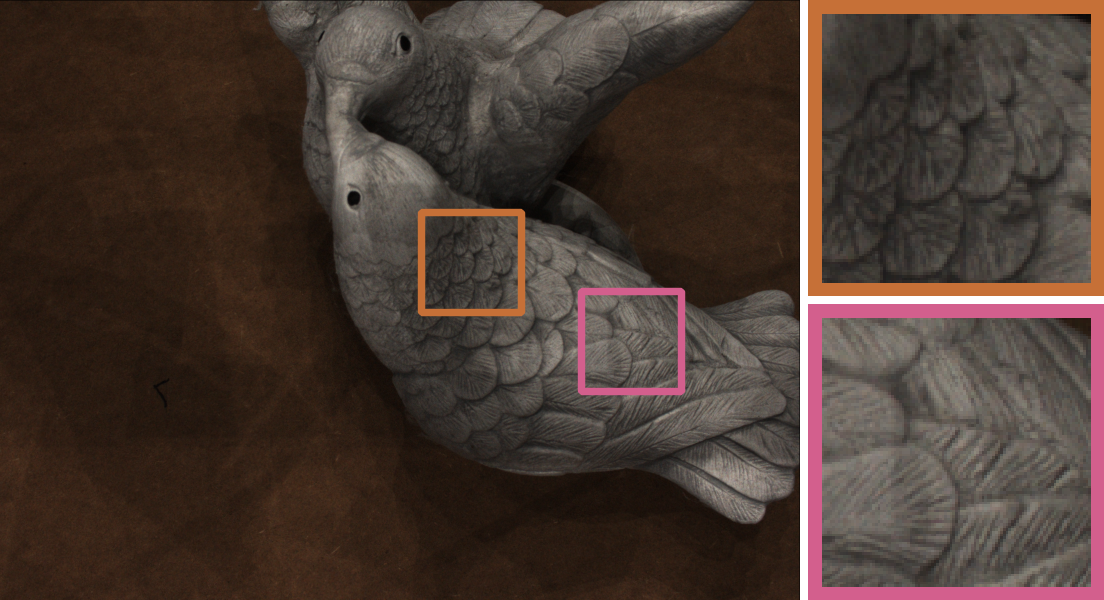}} &
        \includegraphics[trim=100 0 0 0,clip,width=0.33\columnwidth]{{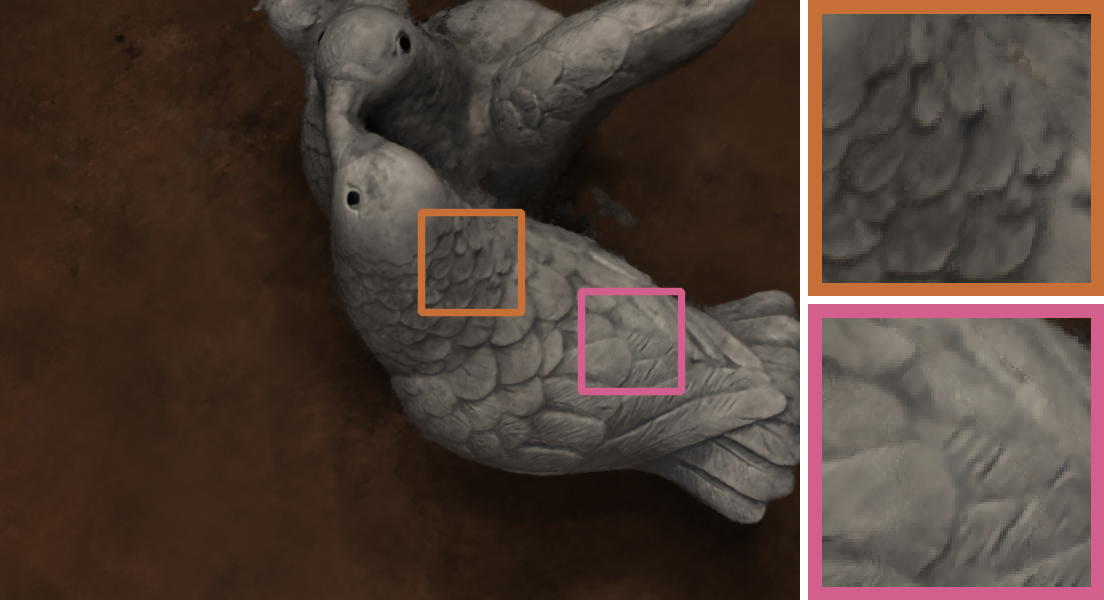}} &
        \includegraphics[trim=100 0 0 0,clip,width=0.33\columnwidth]{{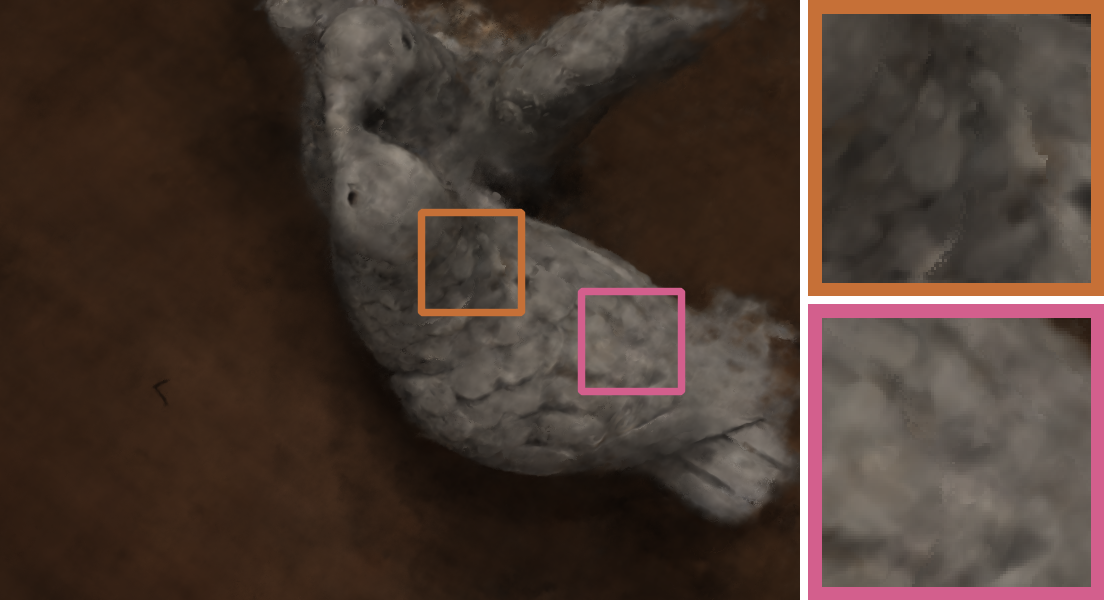}}
        \\
        \includegraphics[trim=0 0 0 0,clip,width=0.33\columnwidth]{{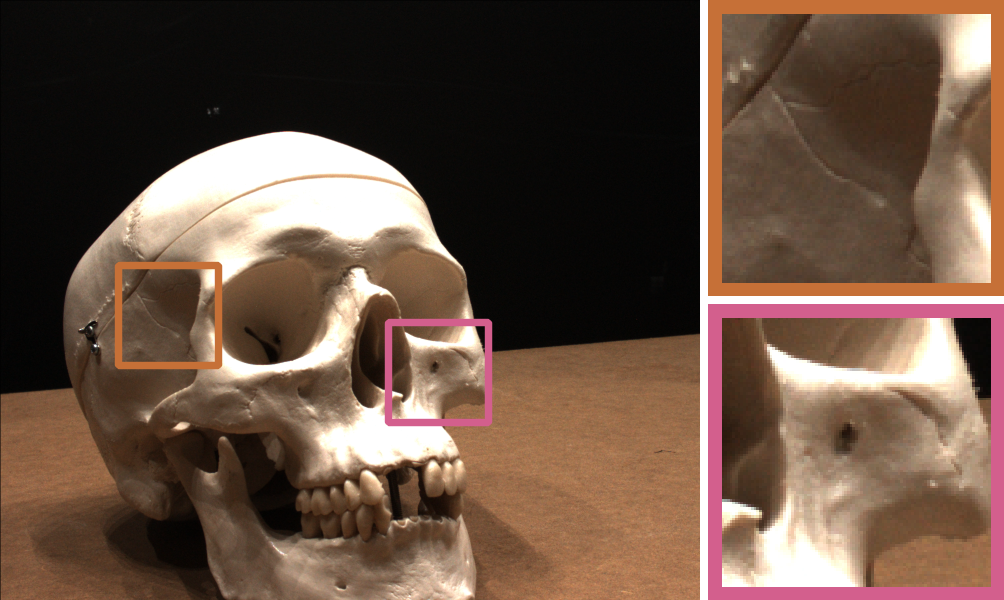}} &
     \includegraphics[trim=0 0 0 0,clip,width=0.33\columnwidth]{{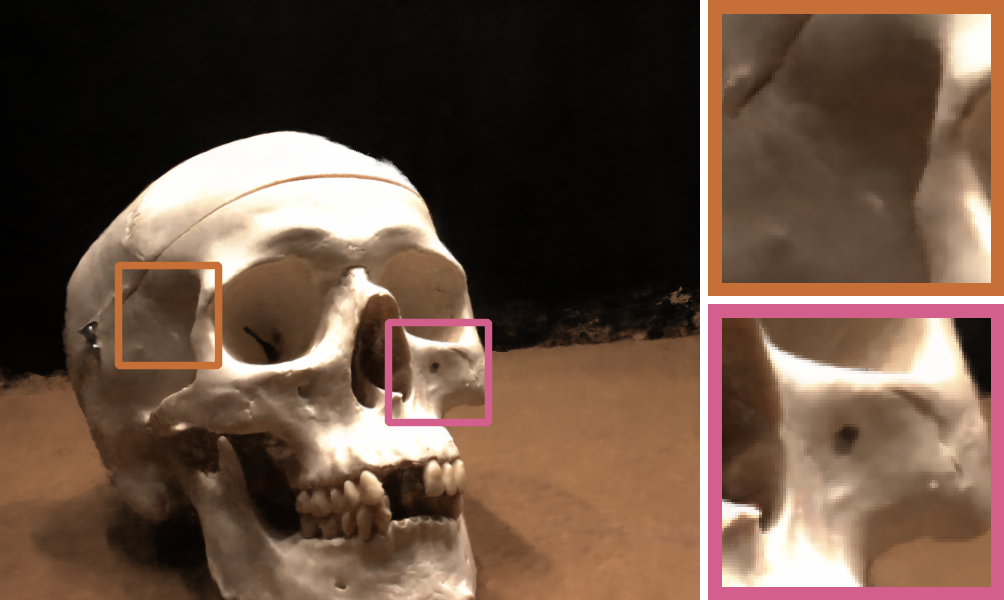}} &
        \includegraphics[trim=0 0 0 0,clip,width=0.33\columnwidth]{{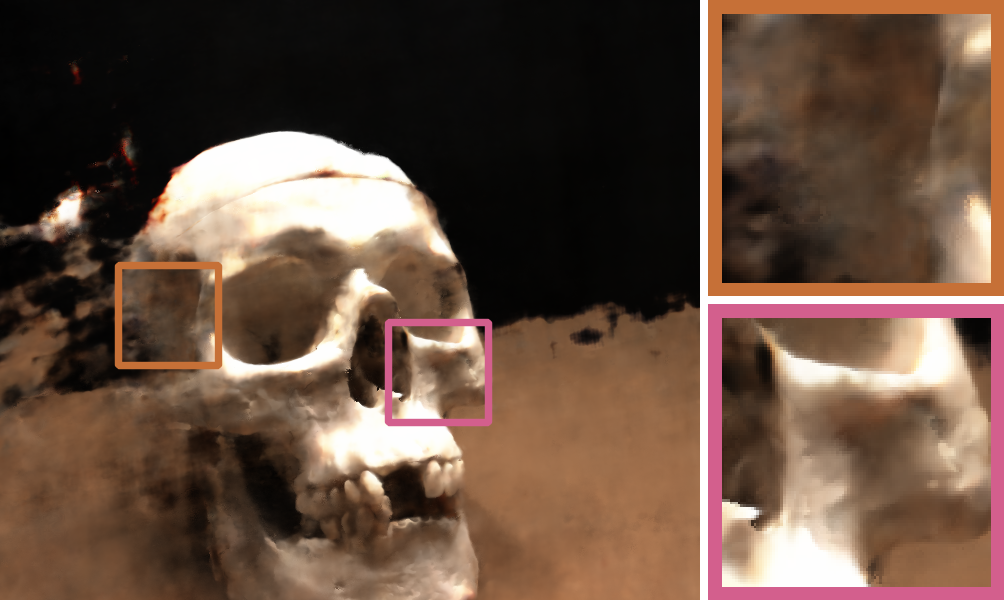}} 
        \\

{\small ground-truth} & {\small \nerf MLP} & {\small vanilla MLP}
     \end{tabular} 
  \captionof{figure}{On DTU scenes~\citep{jensen2014large,Riegler2020FVS}, this figure shows the effect of replacing \nerf's model for the radiance field $\mathbf{c}$ with a vanilla MLP (while keeping the structure of $\sigma$ the same and training both fields from scratch). The vanilla MLP compromises \nerf's ability to generalize to novel views. 
  }\label{fig:implicit_regularize_nerf}
\end{minipage}

\begin{table}
\resizebox{\textwidth}{!}{
\begin{tabular}{l@{\hspace{1em}}c@{\hspace{1em}}c@{\hspace{1em}}c@{\hspace{2em}}c@{\hspace{1em}}c@{\hspace{1em}}c@{\hspace{2em}}c@{\hspace{1em}}c@{\hspace{1em}}c@{\hspace{2em}}ccc}
\toprule 
&\multicolumn{3}{c}{Scan 65} & \multicolumn{3}{c}{Scan 106} & \multicolumn{3}{c}{Scan 118} \\
& $\downarrow$LPIPS & $\uparrow$SSIM & $\uparrow$PSNR & $\downarrow$LPIPS & $\uparrow$SSIM & $\uparrow$PSNR & $\downarrow$LPIPS & $\uparrow$SSIM & $\uparrow$PSNR \\
\nerf MLP & \textbf{0.0258}/\textbf{0.0825} & \textbf{0.988}/\textbf{0.967} & \textbf{33.47}/\textbf{28.41}    & \textbf{0.0641}/\textbf{0.0962} & \textbf{0.978}/\textbf{0.959} & \textbf{35.12}/\textbf{30.65}  & \textbf{0.0380}/\textbf{0.0638} & \textbf{0.987}/\textbf{0.969} & \textbf{37.53}/\textbf{31.82}    \\
Vanilla MLP & 0.0496/0.139 & 0.975/0.937 & 28.55/21.65  & 0.116/0.173 & 0.948/0.915 & 30.29/26.38  & 0.0609/0.094 & 0.979/0.952 & 35.31/29.58 \\
\bottomrule
\end{tabular}
}
\caption{On DTU scenes~\citep{jensen2014large}, replacing \nerf's MLP with a vanilla MLP significantly reduces generalization to novel views. We use the same data split as \cite{Riegler2020FVS}. Numbers on the left are for interpolation, numbers on the right are for extrapolation. They are evaluated on the full images with background masked out.  %
}\label{tab:auxiliary_loss}
\end{table}

\section{Inverted Sphere Parametrization}

\begin{figure}[tb]
\centering
    \begin{tabular}{c@{\hspace{0.02em}}c@{\hspace{0.02em}}c@{\hspace{0.02em}}c@{\hspace{0.02em}}}
        \includegraphics[width=0.5\columnwidth]{{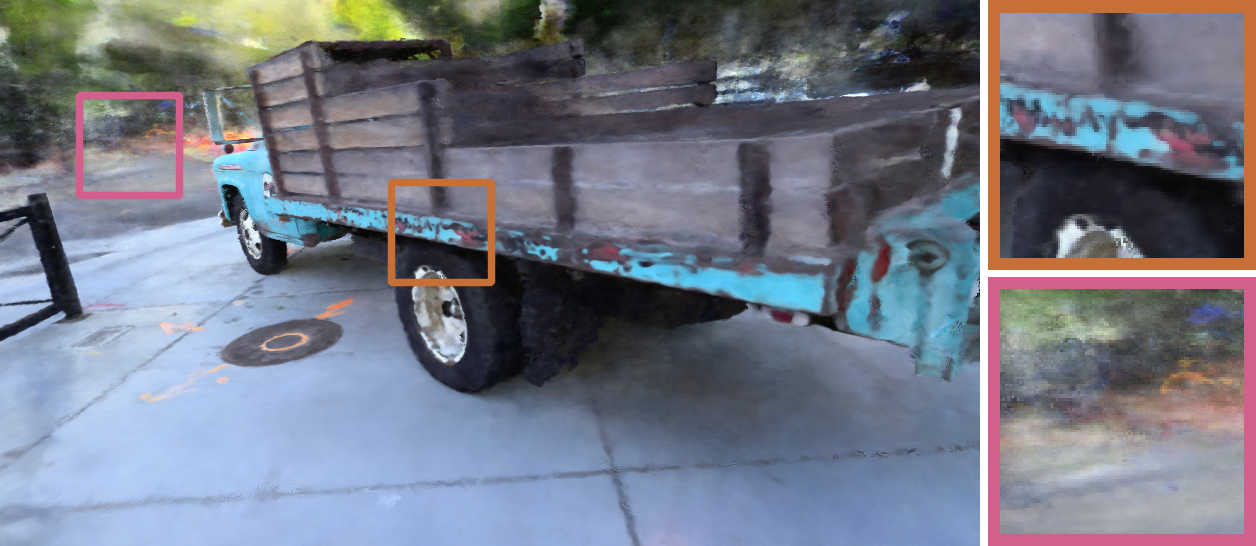}}
        &
        \includegraphics[width=0.5\columnwidth]{{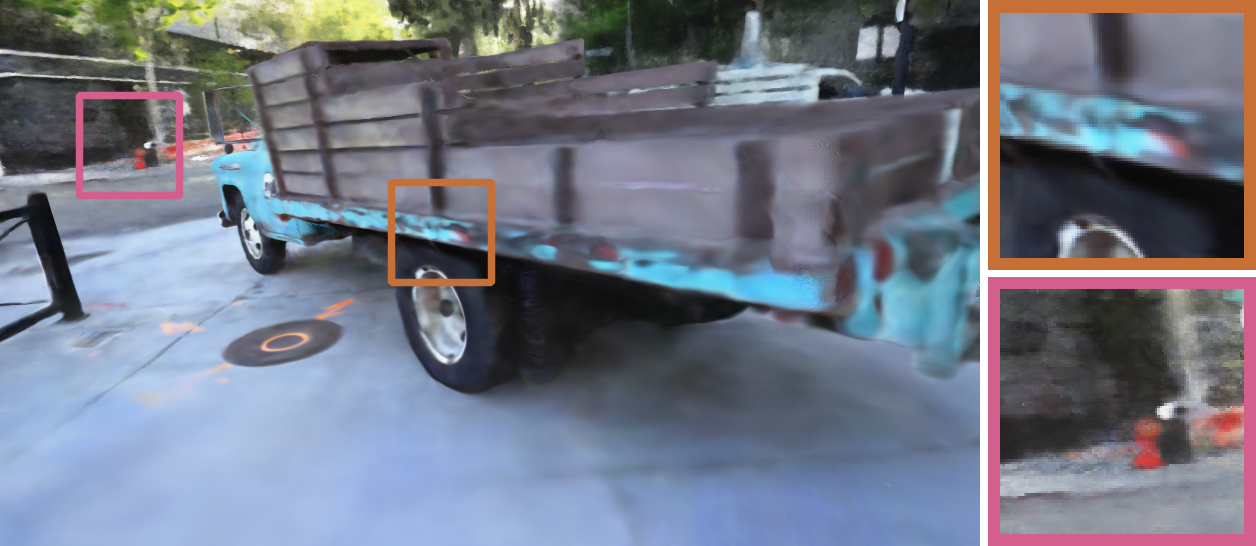}}  \\
{\small (a) bounding volume for the truck only} & {\small (b) bounding volume for the entire scene} 
     \end{tabular} 
  \caption{For 360$^\circ$ captures of unbounded scenes, \nerf's parameterization of space either models only a portion of the scene, leading to significant artifacts in background elements (a), or models the full scene and suffers from an overall loss of detail due to finite sampling resolution (b).
  }\label{fig:truck_tradeoff}
\end{figure}

The volumetric rendering formula in Eq.~\ref{eq:integral} integrates over Euclidean depth. 
When the dynamic range of the true scene depth is small, the integral can be numerically well-approximated with a finite number of samples. 
However, for outdoor, 360$^\circ$ captures centered on nearby objects that also observe the surrounding environment, the dynamic depth range can be extremely large, as the background (buildings, mountains, clouds, etc.) can be arbitrarily far away. 
Such a high dynamic depth range leads to severe resolution issues in \nerf's volumetric scene representation, because to synthesize photo-realistic images, the integral in Eq.~\ref{eq:integral} needs sufficient resolution in both foreground and background areas, which is challenging to achieve by simply sampling points according to a Euclidean parameterization of 3D space. Fig.~\ref{fig:truck_tradeoff} demonstrates this tradeoff between scene coverage and capturing detail. 
In a more restricted scenario where all cameras are forward-facing towards a plane separating the cameras from the scene content, \nerf addresses this resolution issue by projectively mapping a subset of the Euclidean space, i.e., a reference camera's view frustum, to Normalized Device Coordinates (NDC)~\citep{mcreynolds2005advanced}, and integrating in this NDC space. However, this NDC parameterization also fundamentally limits the possible viewpoints, due to its failure to cover the space outside the reference view frustum. 

\begin{wrapfigure}{r}{0.5\textwidth}
 \vspace{-10pt}
  \begin{center}
    \includegraphics[trim=0 0 0 0,clip,width=0.45\textwidth]{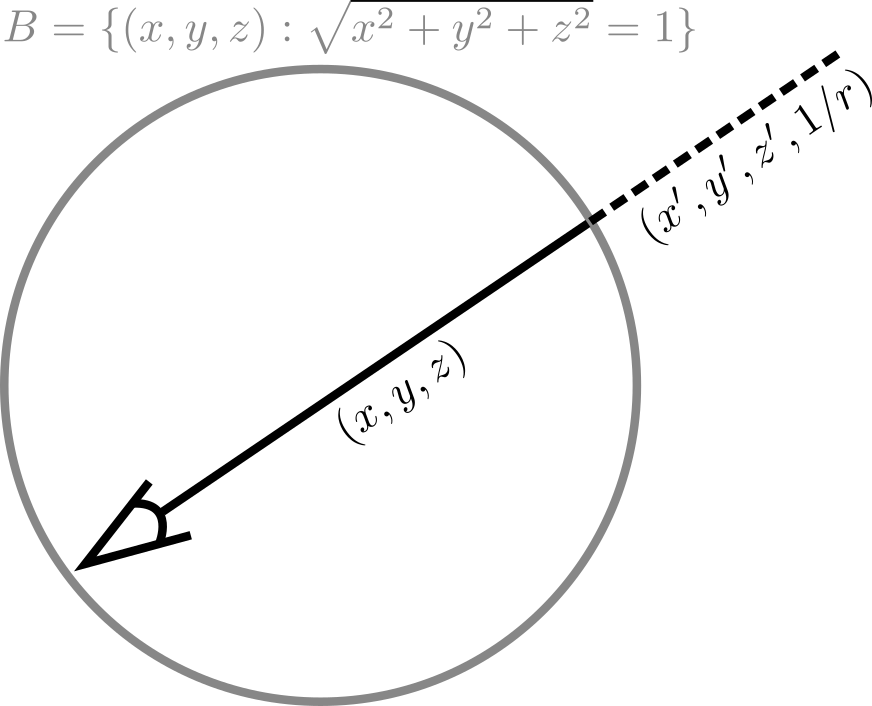}
  \end{center}
\vspace{-10pt}
  \caption{\nerfplusplus applies different parameterizations for scene contents inside and outside the unit sphere.}\label{fig:inverse_sphere}
\end{wrapfigure}
\begin{figure}[b]
\centering
    \begin{tabular}{c@{\hspace{0.02em}}c@{\hspace{0.02em}}c@{\hspace{0.02em}}c@{\hspace{0.02em}}}
        \includegraphics[width=0.33\columnwidth]{{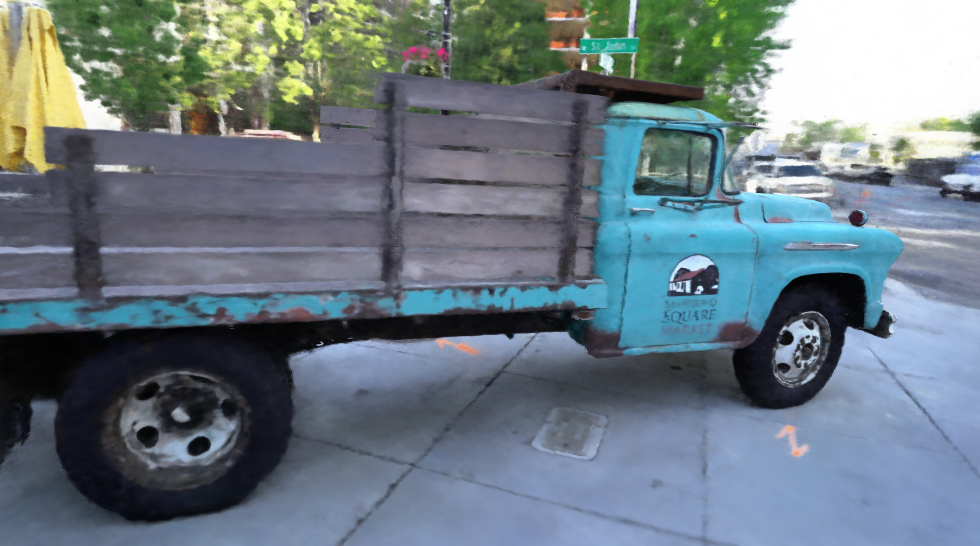}}
        &         \includegraphics[width=0.33\columnwidth]{{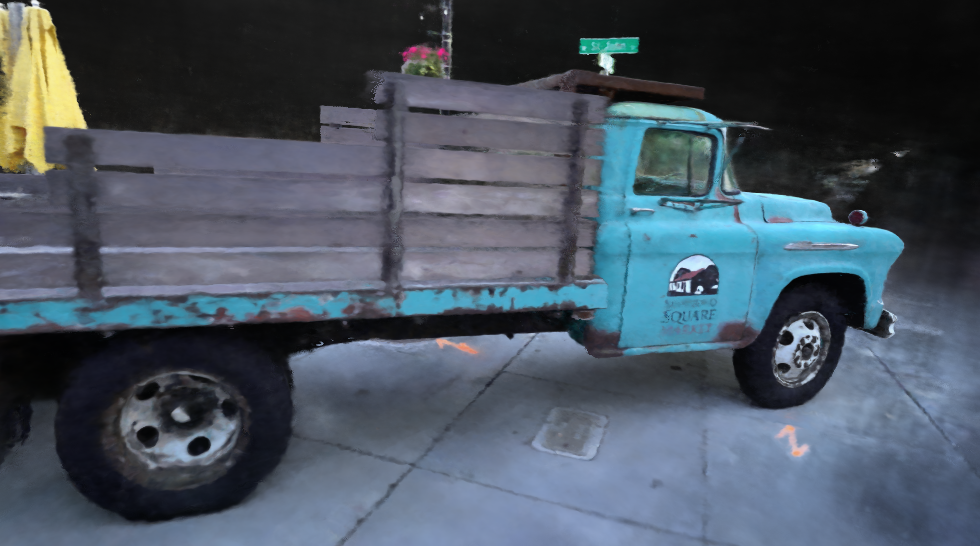}}
        &
        \includegraphics[width=0.33\columnwidth]{{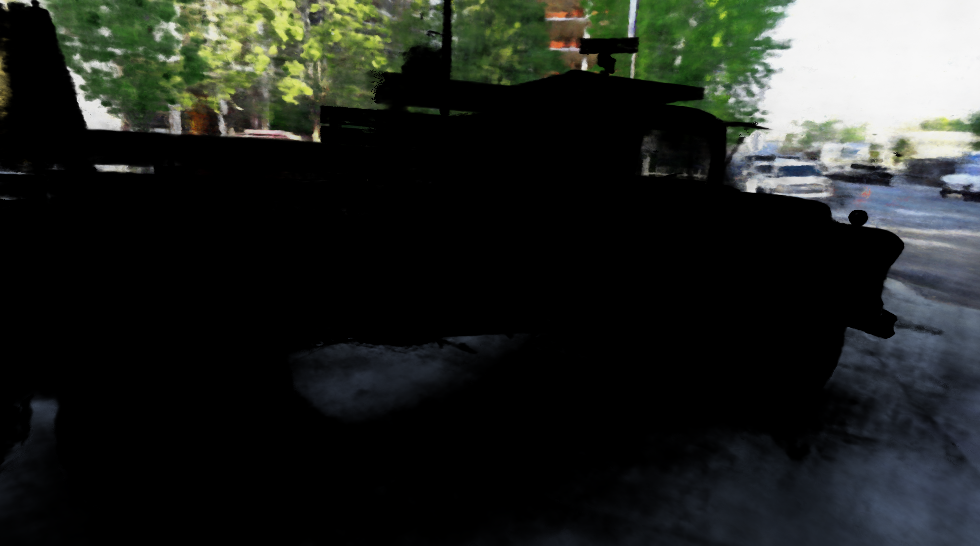}}  \\
{\small (a) \nerfplusplus prediction } & {\small (b) predicted foreground} & {\small (c) predicted background} 
     \end{tabular} 
  \caption{\nerfplusplus separates the modelling of foreground and background. The image synthesized by \nerfplusplus (a) is a composite of the predicted foreground (b) and background (c).
  }\label{fig:truck_separate_fg_bg}
\end{figure}

We address this restriction with an inverted sphere parameterization that facilitates free view synthesis. In our representation, we first partition the scene space into two volumes, an inner unit sphere and an outer volume represented by an inverted sphere covering the complement of the inner volume (see Figure~\ref{fig:inverse_sphere} for an illustration and Figure ~\ref{fig:truck_separate_fg_bg} for a  real-world example of a scene modeled in this way).
The inner volume contains the foreground and all the cameras, while the outer volume contains the remainder of the environment.

These two volumes are modelled with two separate NeRFs.
To render the color for a ray, they are raycast individually, followed by a final compositition. 
No re-parameterization is needed for the inner \nerf, as that part of the scene is nicely bounded. 
For the outer \nerf, we apply an \emph{inverted sphere parametrization}. 

Specifically, a 3D point $(x, y, z),\ \ r=\sqrt{x^2+y^2+z^2}>1$ in the outer volume can be re-parameterized by the quadruple $(x', y', z', 1/r),\ \ x'^2+y'^2+z'^2=1$, where $(x',y',z')$ is a unit vector along the same direction as $(x, y, z)$ representing a direction on the sphere, and $0 < 1/r < 1$ is the inverse radius along this direction specifying the point $r\cdot(x', y', z')$ outside the sphere. 
Unlike Euclidean space where objects can be at unlimited distance from the origin, all the numbers in the re-parametrized quadruple are bounded, i.e., $x', y', z'\in [-1, 1], 1/r\in [0,1]$. This not only improves numeric stability, but also respects the fact that farther objects should get less resolution.
We can directly raycast this 4D bounded volume (only 3 degrees of freedom) to render a camera ray's color.
Note that the composite of foreground and background is equivalent to breaking the integral in Eq.~\ref{eq:integral} into two parts, integration inside the inner and outer volumes.
In particular, consider that the ray $\mathbf{r}=\mathbf{o}+t\mathbf{d}$ is partitioned into two segments by the unit sphere: in the first, $t\in(0, t')$ is inside the sphere; in the second, $t\in(t', \infty)$ is outside the sphere. We can rewrite the volumetric rendering integral in Eq.~\ref{eq:integral} as
\begin{align}
    \mathbf{C}(\mathbf{r})
    &=\underbrace{\int_{t=0}^{t'} \sigma(\mathbf{o}+t\mathbf{d})\cdot \mathbf{c}(\mathbf{o}+t\mathbf{d}, \mathbf{d})\cdot e^{-\int_{s=0}^{t}\sigma(\mathbf{o}+s\mathbf{d})ds}dt}_{\text{(i)}} \nonumber\\
    &\quad +\underbrace{e^{-\int_{s=0}^{t'}\sigma(\mathbf{o}+s\mathbf{d})ds}}_{\text{(ii)}}\cdot\underbrace{\int_{t=t'}^{\infty} \sigma(\mathbf{o}+t\mathbf{d})\cdot \mathbf{c}(\mathbf{o}+t\mathbf{d}, \mathbf{d})\cdot e^{-\int_{s=t'}^{t}\sigma(\mathbf{o}+s\mathbf{d})ds}dt}_{\text{(iii)}}.
\label{eq:integral_rewrite}
\end{align}
Terms (i) and (ii) are computed in Euclidean space, while term (iii) is computed in inverted sphere space with $\frac{1}{r}$ as the integration variable. In other words, we use $\sigma_{\mathit{in}}(\mathbf{o}+t\mathbf{d}), \mathbf{c}_{\mathit{in}}(\mathbf{o}+t\mathbf{d}, \mathbf{d})$ in terms (i) and (ii), and $\sigma_{\mathit{out}}(x',y',z',1/r), \mathbf{c}_{\mathit{out}}(x',y',z',1/r, \mathbf{d})$ in term (iii).

In order to compute term (iii) for the ray $\mathbf{r} = \mathbf{o}+t\mathbf{d}$, we first need to be able to evaluate $\sigma_{\mathit{out}},\mathbf{c}_{\mathit{out}}$ at any $1/r$; in other words, we need a way to compute $(x',y',z')$ corresponding to a given $1/r$, so that $\sigma_{\mathit{out}}, \mathbf{c}_{\mathit{out}}$ can take $(x',y',z',1/r)$ as input. This can be achieved as follows.
\begin{wrapfigure}{r}{0.5\textwidth}
 \vspace{-10pt}
  \begin{center}
    \includegraphics[trim=0 0 0 0,clip,width=0.45\textwidth]{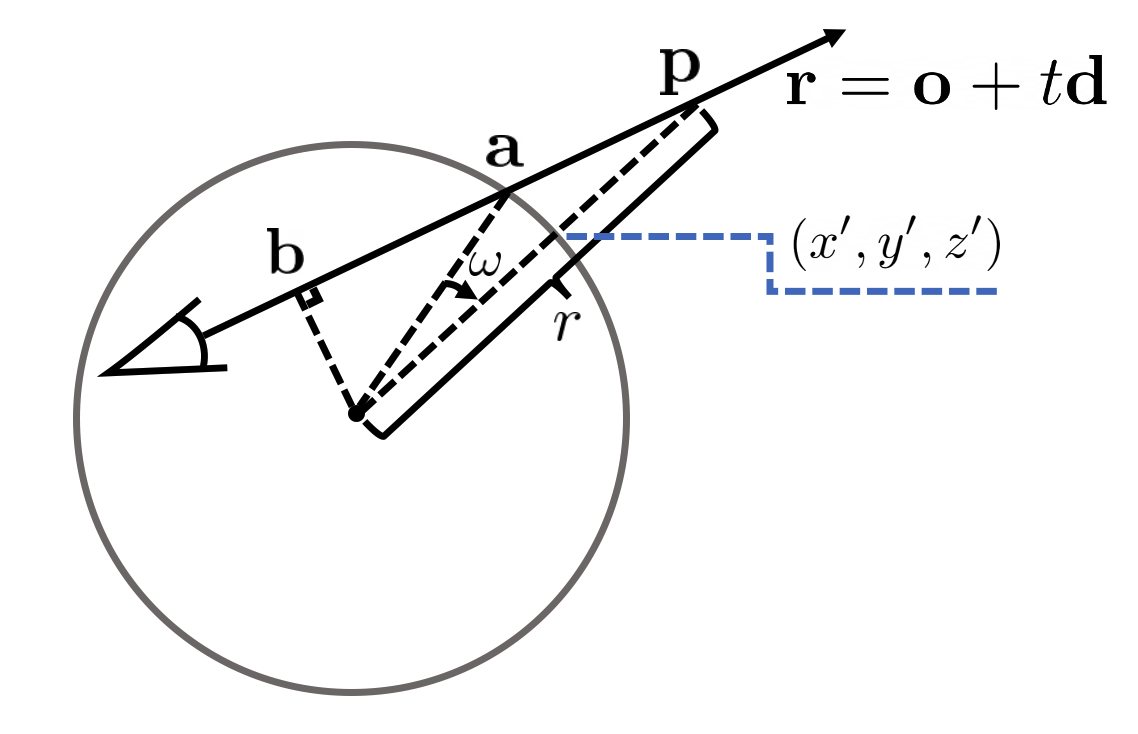}
  \end{center}
\vspace{-10pt}
  \caption{{\small Diagram illustrating the derivation of $(x',y',z')$ for point $\mathbf{p}$ with known $1/r$ in the inverted sphere parameterization.}}\label{fig:inverse_sphere_int}
\end{wrapfigure}
As shown in Fig.~\ref{fig:inverse_sphere_int}, let the ray intersect the unit sphere at point $\mathbf{a}$, and the midpoint of the chord aligning with the ray be the point $\mathbf{b}$. Point ${\mathbf{a}=\mathbf{o}+t_a\mathbf{d}}$ is computed by solving $\vert \mathbf{o}+t_a\mathbf{d}\vert=1$, while $\mathbf{b}=\mathbf{o}+t_b\mathbf{d}$ is acquired by solving ${\mathbf{d}^T(\mathbf{o}+t_b\mathbf{d})=0}$. Then to get $(x',y',z')$ given $1/r$, we can rotate the vector $\mathbf{a}$ along the vector $\mathbf{b}\times \mathbf{d}$ by the angle $\omega=\arcsin\vert \mathbf{b}\vert-\arcsin(\vert \mathbf{b}\vert\cdot\frac{1}{r})$.
Once we can evaluate $\sigma_{\mathit{out}},\mathbf{c}_{\mathit{out}}$ at any $1/r$, we simply sample a finite number of points from the interval $[0, 1]$ to compute term (iii).

The inverted sphere parameterization for the outer volume has an intuitive physical explanation. 
It can be viewed in terms of a virtual camera whose image plane is the unit sphere at the scene origin.
Hence, the 3D point $(x, y, z)$ is projected to the pixel $(x', y', z')$ on the image plane, while the term $1/r\in (0, 1)$ serves as the (inverse) depth, or disparity, of the point. 
From this perspective, the NDC parameterization suitable only for forward-facing capture is related to our representation, as it uses a virtual pinhole camera rather than a spherical projection surface. In this sense, our inverted sphere parameterization is related to the concept of \emph{multi-sphere images} (a scene representation consisting of nested concentric spheres sampled according to inverse depth from the sphere centers) proposed in recent work on view synthesis~\citep{attal2020matryodshka,broxton2020immersive}.

\section{Experiments}
\label{sec:experiment}

We validate \nerfplusplus and compare it with \nerf on two real-world datasets captured with hand-held cameras: Tanks and Temples (T\&T)~\citep{Knapitsch2017} and the Light Field (LF) dataset of~\citet{yucer2016efficient}. 
We report PSNR, SSIM, and LPIPS~\citep{zhang2018perceptual} as our quantitative metrics for measuring the quality of synthesized test images.

\textbf{T\&T dataset.} We use the training/testing images and SfM poses provided by \cite{Riegler2020FVS}. This dataset consists of hand-held 360$^\circ$ captures of four large-scale scenes; the camera poses are estimated by COLMAP SfM~\citep{schoenberger2016sfm}. For \nerf, we normalize the scene such that all cameras are inside the sphere of radius $\frac{1}{8}$. This normalization ensures that the unit sphere covers the majority of the background scene content (although some background geometry still lies outside the bounding unit sphere). To numerically compute the volumetric rendering integral for each camera ray, we uniformly sample points from the ray origin to its intersection with the unit sphere. Since \nerfplusplus ray-casts both inner and outer volumes and hence uses twice the number of samples per camera ray compared to a single volume, we also double the number of samples used by \nerf for the sake of fairness. Specifically, \nerf's coarse-level MLP uses 128 uniform samples, while the fine-level MLP uses 256 additional importance samples.  Under this hyper-parameter setting, \nerf has roughly the same computation cost and GPU memory footprint during training and testing as \nerfplusplus. We randomly sample 2048 camera rays at each training iteration, and train \nerf and \nerfplusplus for 250k iterations with a learning rate of 5e-4.

\textbf{LF dataset.} We use four scenes from the LF dataset: Africa, Basket, Ship, and Torch. Each scene is densely covered by 2k--4k hand-held captured images, and camera parameters are recovered using SfM.  We construct a sparse surrounding capture by temporally subsampling the images by a factor of 40. In particular, the training images are frames 0, 40, 80, ..., and the testing images are frames 20, 60, 100, ... .
The scene normalization method, number of samples per camera ray, batch size, training iterations, and learning rate are the same as in the T\&T dataset.

\begin{figure}[t]
\centering
    \begin{tabular}{c@{\hspace{0.02em}}c@{\hspace{0.02em}}}
        \includegraphics[width=0.5\columnwidth]{{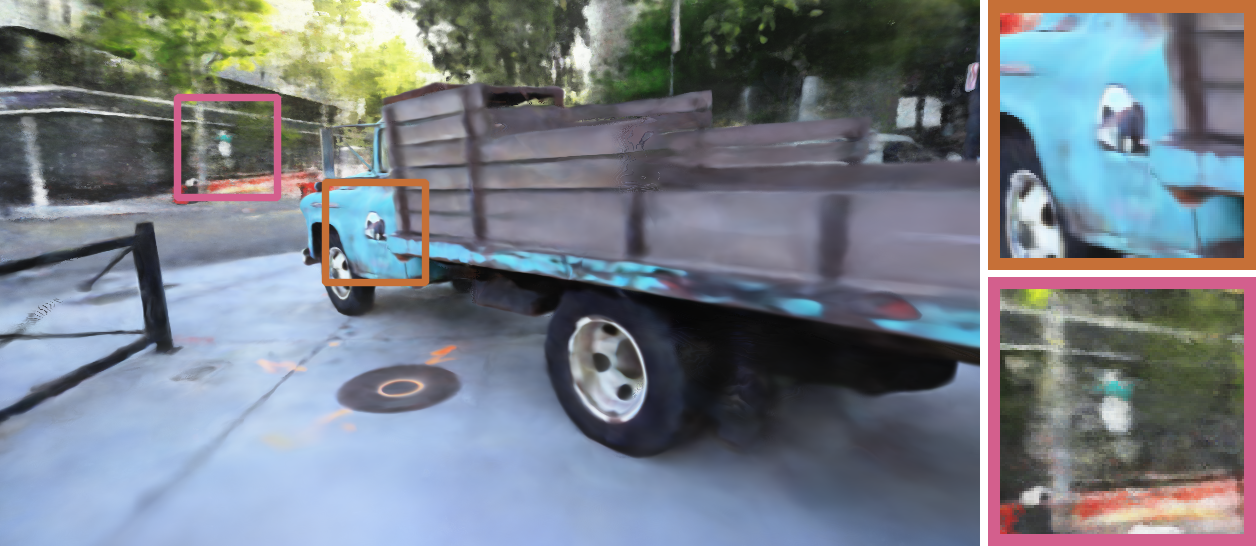}}  & 
        \includegraphics[width=0.5\columnwidth]{{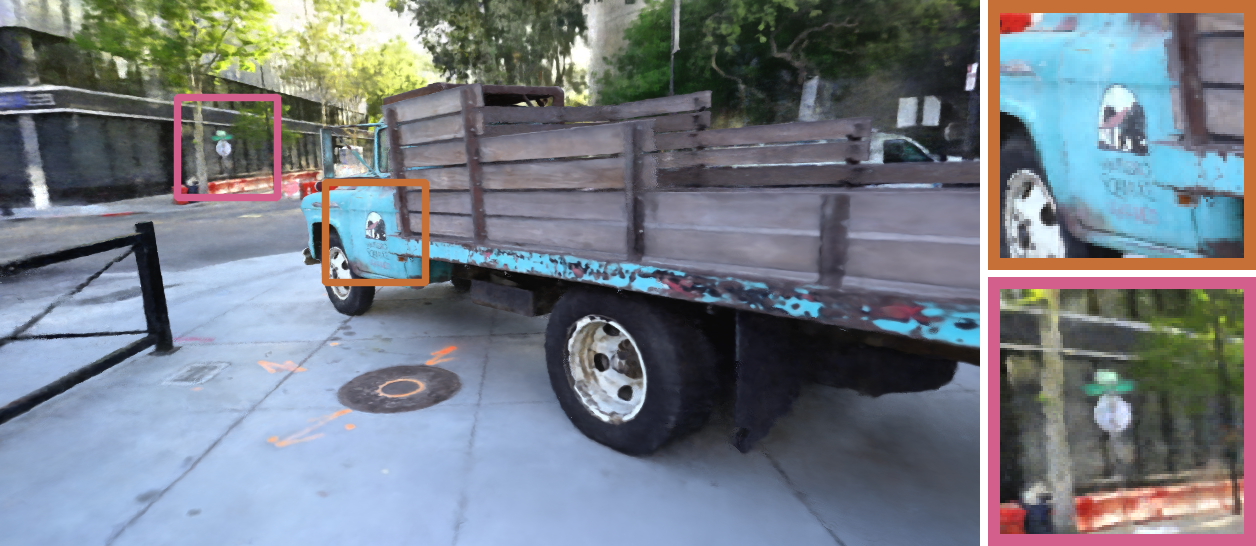}} \\
        \includegraphics[width=0.5\columnwidth]{{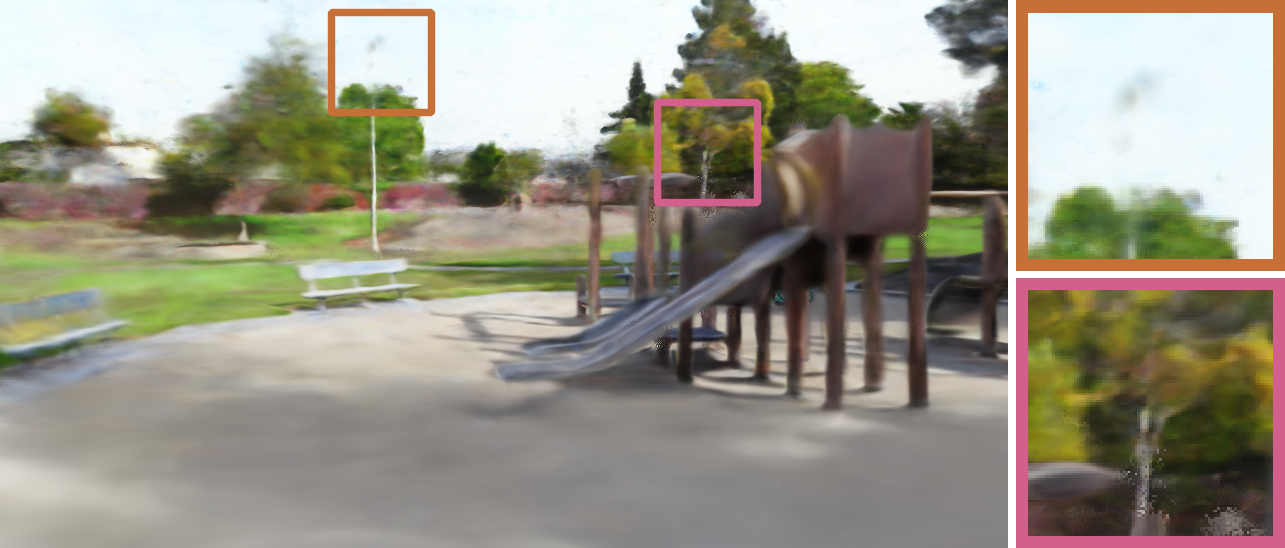}}
        &
        \includegraphics[width=0.5\columnwidth]{{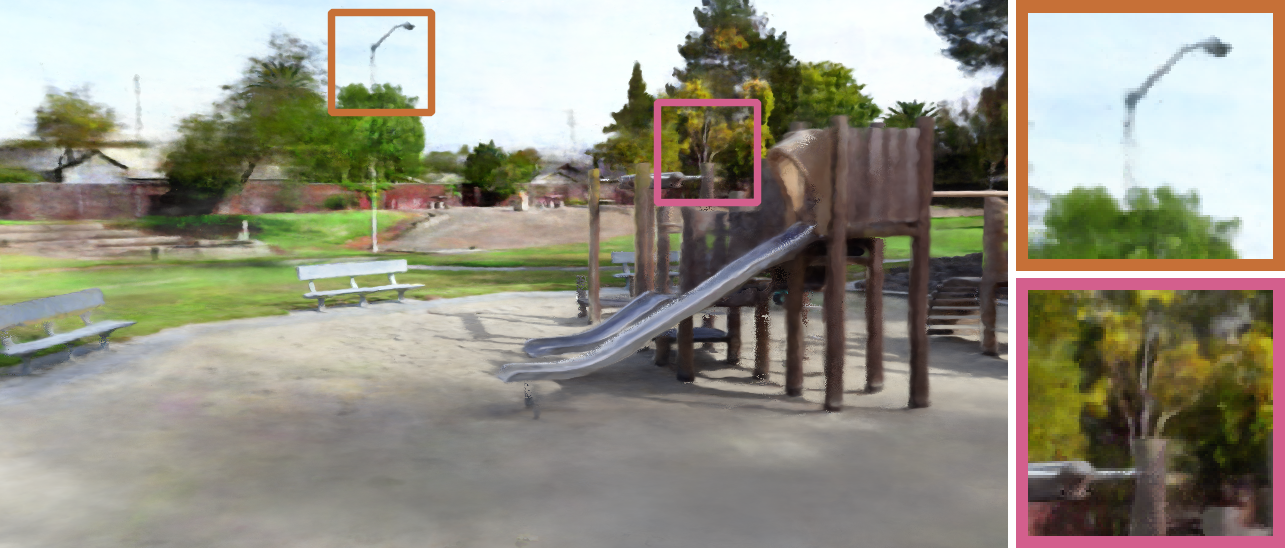}}  \\
       \includegraphics[width=0.5\columnwidth]{{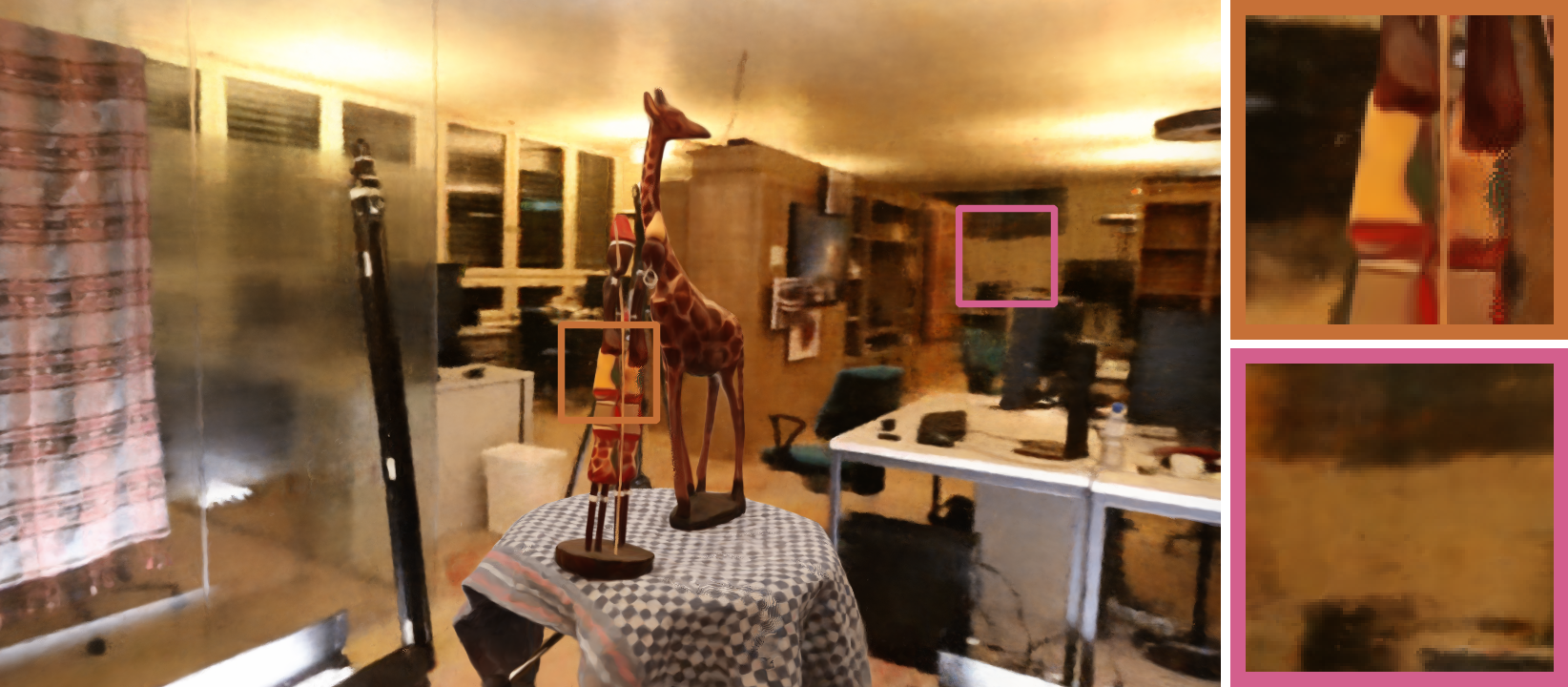}}
        &
        \includegraphics[width=0.5\columnwidth]{{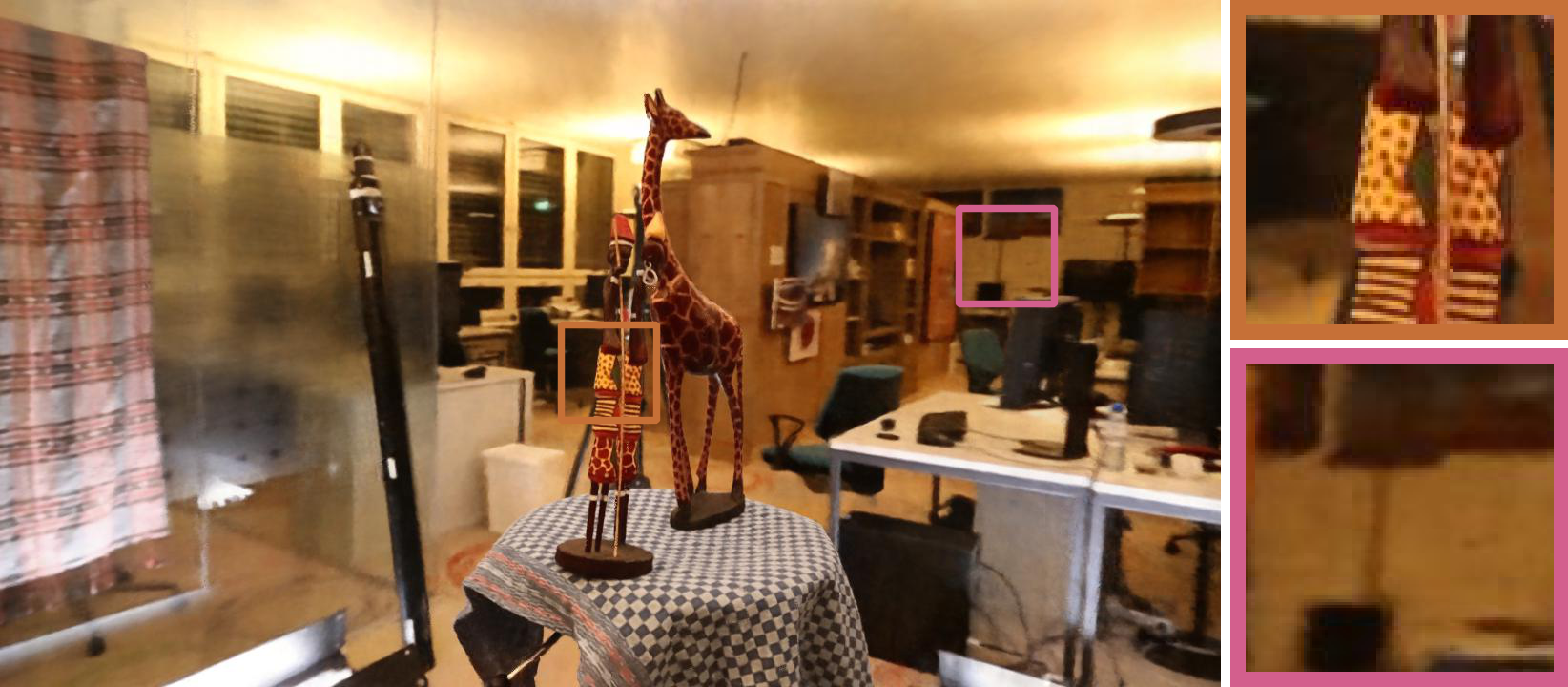}}  \\
     \includegraphics[width=0.5\columnwidth]{{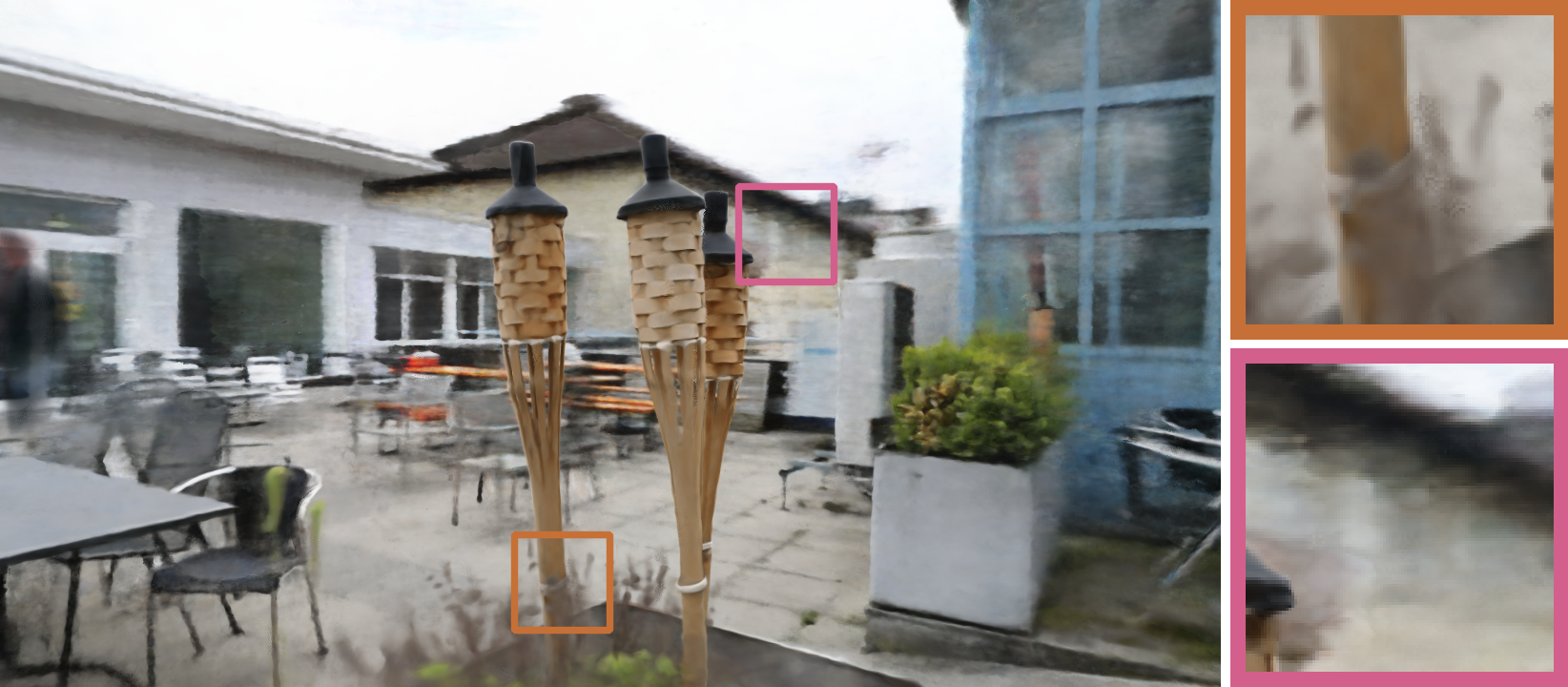}}
        &
        \includegraphics[width=0.5\columnwidth]{{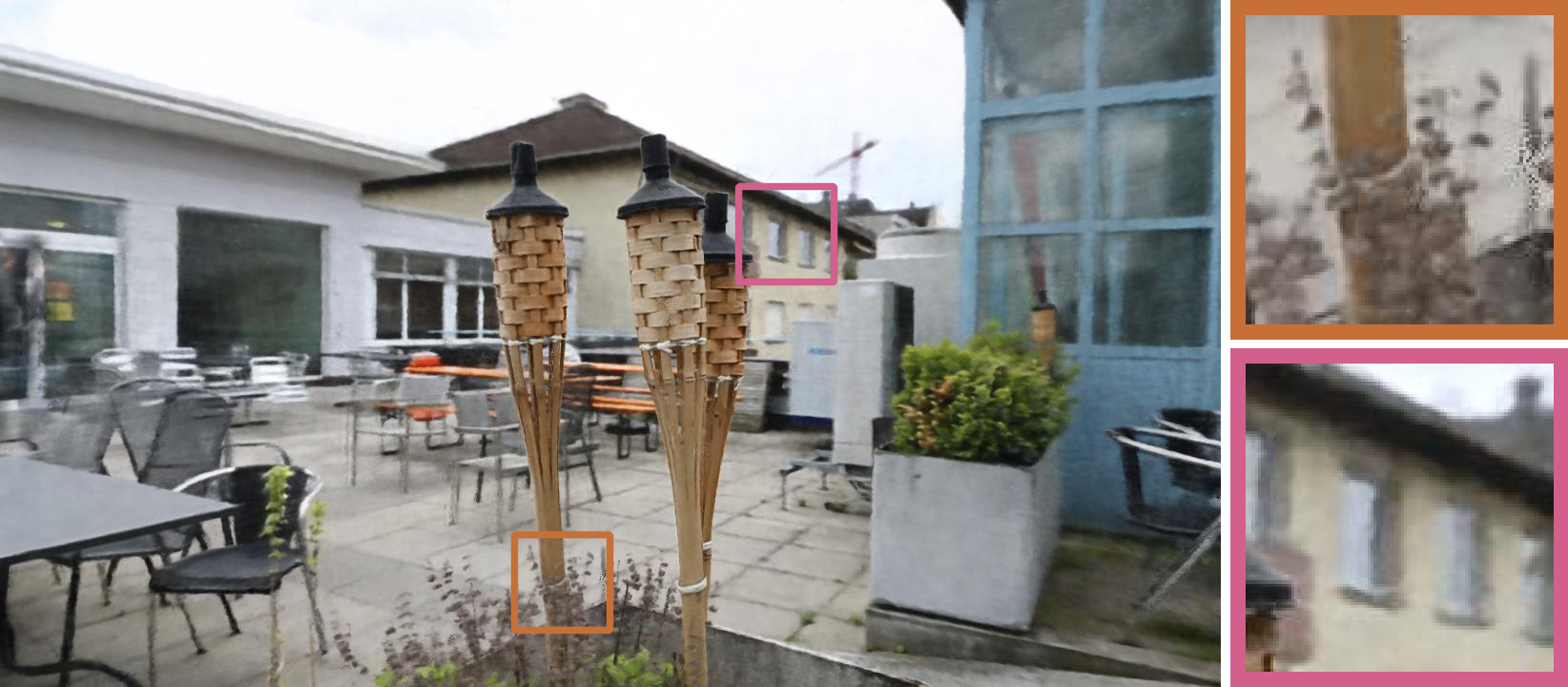}}  \\
{\small \nerf } & {\small \nerfplusplus}  \\
     \end{tabular} 
  \caption{We qualitatively compare \nerfplusplus with \nerf on two T\&T scenes (Truck, Playgroud) and two LF scenes (Africa, Torch). \nerfplusplus produces sharper images than \nerf and is able to better represent both the foreground and the background.}\label{fig:real_compare_alg}
\end{figure}

\textbf{Results.} As shown in Table~\ref{tab:main_results}, \nerfplusplus significantly outperforms \nerf in challenging scenarios involving 360$^\circ$ captures of objects within large-scale unbounded scenes. In Figure~\ref{fig:real_compare_alg}, we can see that images synthesized by \nerfplusplus have significantly higher fidelity.

\begin{table}
\resizebox{\textwidth}{!}{
\begin{tabular}{l@{\hspace{1em}}c@{\hspace{1em}}c@{\hspace{1em}}c@{\hspace{2em}}c@{\hspace{1em}}c@{\hspace{1em}}c@{\hspace{2em}}c@{\hspace{1em}}c@{\hspace{1em}}c@{\hspace{2em}}ccc}
\toprule 
&\multicolumn{3}{c}{Truck} & \multicolumn{3}{c}{Train} & \multicolumn{3}{c}{M60} & \multicolumn{3}{c}{Playground} \\
& $\downarrow$LPIPS & $\uparrow$SSIM & $\uparrow$PSNR & $\downarrow$LPIPS & $\uparrow$SSIM & $\uparrow$PSNR & $\downarrow$LPIPS & $\uparrow$SSIM & $\uparrow$PSNR & $\downarrow$LPIPS & $\uparrow$SSIM & $\uparrow$PSNR \\
\nerf & 0.513 & 0.747 & 20.85    & 0.651 & 0.635 & 16.64  & 0.602 & 0.702 & 16.86 &   0.529 & 0.765 & 21.55\\
\nerfplusplus  & \textbf{0.298}& \textbf{0.823} & 
\textbf{22.77}   & \textbf{0.523} & \textbf{0.672} & \textbf{17.17}  & \textbf{0.435} & \textbf{0.738} & \textbf{17.88} & \textbf{0.391} & \textbf{0.799} & \textbf{22.37}\\
\midrule
&\multicolumn{3}{c}{Africa} & \multicolumn{3}{c}{Basket} & \multicolumn{3}{c}{Torch} & \multicolumn{3}{c}{Ship} \\
& $\downarrow$LPIPS & $\uparrow$SSIM & $\uparrow$PSNR & $\downarrow$LPIPS & $\uparrow$SSIM & $\uparrow$PSNR & $\downarrow$LPIPS & $\uparrow$SSIM & $\uparrow$PSNR & $\downarrow$LPIPS & $\uparrow$SSIM & $\uparrow$PSNR \\
\nerf & 0.217 & 0.894 & 26.16 & 0.377 & 0.805 & 20.83 & 0.347 & 0.811 & 22.81 & 0.372 & 0.801 & 23.24 \\
\nerfplusplus & \textbf{0.163} & \textbf{0.923} &\textbf{27.41} & \textbf{0.254} & \textbf{0.884} & \textbf{21.84}  & \textbf{0.226} & \textbf{0.867} & \textbf{24.68} & \textbf{0.241} & \textbf{0.867} & \textbf{25.35} \\
\bottomrule
\end{tabular}
}
\caption{We compare \nerfplusplus with \nerf on four Tanks and Temples scenes: Truck, Train, M60, and Playground, as well as four LF scenes: Africa, Torch, Ship, and Basket. \nerfplusplus consistently outperforms \nerf in all metrics.}\label{tab:main_results}
\end{table}

\section{Open challenges}
\nerfplusplus improves the parameterization of unbounded scenes in which both the foreground and the background need to be faithfully represented for photorealism. However, there remain a number of open challenges. 
First, the training and testing of \nerf and \nerfplusplus on a single large-scale scene is quite time-consuming and memory-intensive. Training \nerfplusplus on a node with 4 RTX 2080 Ti GPUs takes $\sim$24 hours. Rendering a single 1280x720 image on one such GPU takes $\sim$30 seconds at test time.
\cite{liu2020neural} have sped up the inference, but rendering is still far from real-time.
Second, small camera calibration errors may impede photorealistic synthesis. Robust loss functions, such as the contextual loss~\citep{mechrez2018contextual}, could be applied. 
Third, photometric effects such as auto-exposure and vignetting can also be taken into account to increase image fidelity. This line of investigation is related to the lighting changes addressed in the orthogonal work of \cite{martin2020nerf}.

\bibliography{iclr2020_conference}
\bibliographystyle{iclr2020_conference}

\end{document}